\documentclass[letterpaper]{article} 
\usepackage{aaai24}  
\usepackage{times}  
\usepackage{helvet}  
\usepackage{courier}  
\usepackage[hyphens]{url}  
\usepackage{graphicx} 
\usepackage{natbib}  
\usepackage{caption} 
\usepackage{algorithm}
\usepackage{algorithmic}

\usepackage{newfloat}
\usepackage{listings}
\DeclareCaptionStyle{ruled}{labelfont=normalfont,labelsep=colon,strut=off} 
\floatstyle{ruled}
\newfloat{listing}{tb}{lst}{}
\floatname{listing}{Listing}
\title{Dual-Perspective Knowledge Enrichment for
Semi-Supervised \\ 3D Object Detection}
\author{
    Yucheng Han\textsuperscript{\rm 1}, Na Zhao\textsuperscript{\rm 2}\thanks{Corresponding author.}, Weiling Chen\textsuperscript{\rm 3}, Keng Teck Ma\textsuperscript{\rm 3}, Hanwang Zhang\textsuperscript{\rm 1}
}
\affiliations{
    \textsuperscript{\rm 1}Nanyang Technological University\\
    \textsuperscript{\rm 2}Singapore University of Technology and Design\\
    \textsuperscript{\rm 3}Hyundai Motor Group Innovation Center in Singapore\\

    yucheng002@ntu.edu.sg,~na\_zhao@sutd.edu.sg,~\{weiling.chen,~kengteck.ma\}@hmgics.com,~hanwangzhang@ntu.edu.sg
}

\usepackage{bibentry}

\usepackage{mathtools}
\usepackage{subcaption}
\usepackage{bm}
\usepackage{gensymb}
\usepackage{amsmath}
\usepackage{amssymb}
\usepackage{bbold}
\usepackage{booktabs}

\usepackage{multirow}

\begin{document}

\maketitle

\begin{abstract}

Semi-supervised 3D object detection is a promising yet under-explored direction to reduce data annotation costs, especially for cluttered indoor scenes. A few prior works, such as SESS and 3DIoUMatch, attempt to solve this task by utilizing a teacher model to generate pseudo-labels for unlabeled samples. However, the availability of unlabeled samples in the 3D domain is relatively limited compared to its 2D counterpart due to the greater effort required to collect 3D data. Moreover, the loose consistency regularization in SESS and restricted pseudo-label selection strategy in 3DIoUMatch lead to either low-quality supervision or a limited amount of pseudo labels. To address these issues, we present a novel Dual-Perspective Knowledge Enrichment approach named DPKE for semi-supervised 3D object detection. Our DPKE enriches the knowledge of limited training data, particularly unlabeled data, from two perspectives: data-perspective and feature-perspective. Specifically, from the data-perspective, we propose a class-probabilistic data augmentation method that augments the input data with additional instances based on the varying distribution of class probabilities. Our DPKE achieves feature-perspective knowledge enrichment by designing a geometry-aware feature matching method that regularizes feature-level similarity between object proposals from the student and teacher models. Extensive experiments on the two benchmark datasets demonstrate that our DPKE achieves superior performance over existing state-of-the-art approaches under various label ratio conditions. The source code will be made available to the public.
\end{abstract}

\section{Introduction}\label{sec:intro}

\begin{figure}[t]
    \centering
    \begin{minipage}[t]{.65\linewidth}
        \centering
        \includegraphics[width=.95\linewidth]{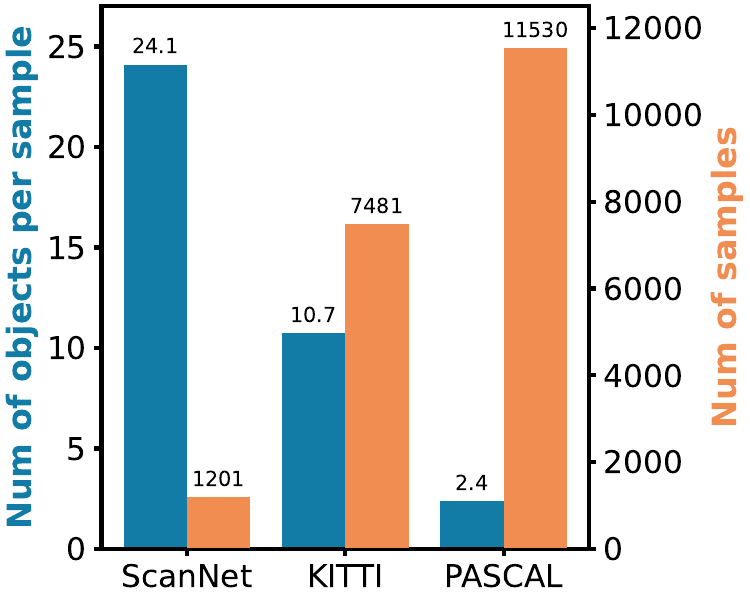}
    \end{minipage}
    \caption{\small{\textbf{The dataset statistics}. The orange bars represent the number of samples in the corresponding dataset, while the blue bars represent the number of objects per sample. The 3D indoor dataset (ScanNet) contains much more objects per scene than the 3D outdoor dataset (KITTI) or the 2D dataset (Pascal).}}
    \label{fig:Fig1}
\end{figure}

3D object detection, which can localize and categorize objects of interest in a 3D scene, is a crucial technology for numerous applications, including domestic robotics, autonomous driving, and augmented reality. Despite the effectiveness of modern 3D object detection approaches \cite{yang2023swin3d,wang2022cagroup3d,rukhovich2023tr3d,zhang2022glenet,xu2022behind,zheng2021se}, their performance heavily relies on the availability of a large amount of well-annotated 3D data. However, compared to its 2D counterpart, annotating 3D data is more time-consuming and expensive. To be more specific, the process of annotating a single image takes approximately 0.7 minutes \cite{papadopoulos2016we}, whereas annotating an outdoor 3D scene can require up to 1.9 minutes \cite{lee2018leveraging}. The task becomes even more arduous in the case of cluttered indoor 3D scenarios, such as annotating a single scan in the 3D indoor benchmark dataset ScanNet \cite{dai2017scannet}, which can take up to 22.3 minutes. The excessively high annotation costs make the scale of 3D indoor datasets significantly smaller than that of 3D outdoor datasets and 2D datasets as shown in Figure~\ref{fig:Fig1}. In order to alleviate the burden of labor-intensive data annotation, our research delves into semi-supervised 3D object detection, which aims to achieve comparable performance to fully supervised approaches while reducing the amount of required annotated data by exploring knowledge from unlabeled data.


\begin{figure*}[t]
    \centering
    \includegraphics[width=.75\textwidth]{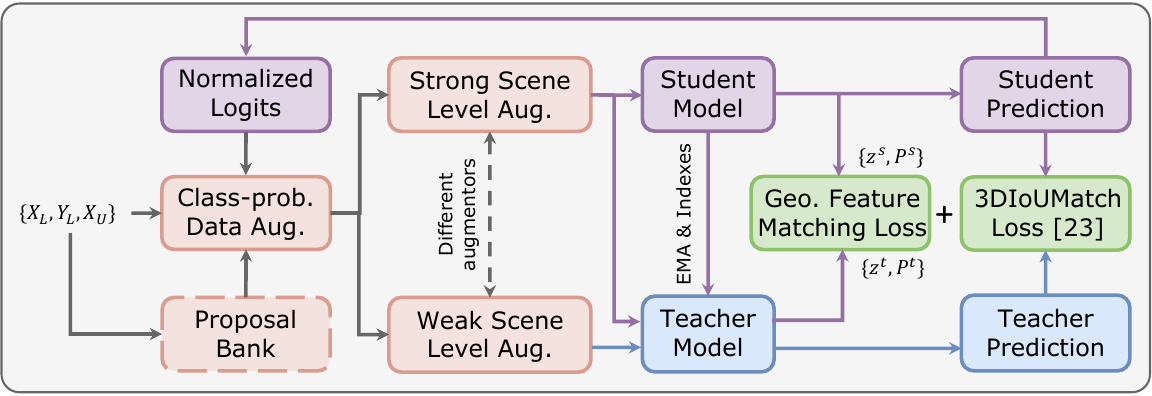}
    \caption{\small{\textbf{The overall framework of our proposed DPKE}. Before the original scene-level augmentation operator, we introduce a new data augmentation module based on the normalized logits of the student model to increase the diversity of the input data.
    The student model is updated using the loss function combined by 3DIoUMatch Loss \cite{wang20213dioumatch} and our geometry-aware feature matching loss, which exploits knowledge from teacher predictions with lower confidence.
    }}
    \label{fig:pipeline}
\end{figure*}

A few prior works \cite{zhao2020sess, wang20213dioumatch, chen2023class} on semi-supervised 3D object detection adopt the Mean Teacher paradigm \cite{tarvainen2017mean}, which involves a student and a teacher model, with the teacher model providing supervision to the student outputs of the unlabeled data. Specifically, SESS~\cite{zhao2020sess} considers all outputs of the teacher model as pseudo labels and enforces the consistency regularization between two model outputs in terms of center, size, and class. As a result, SESS tends to produce imprecise pseudo labels that exhibit high recall but poor quality. In contrast, methods like 3DIoUMatch \cite{wang20213dioumatch} apply a strict threshold strategy to select high-quality pseudo labels, which leads to a lower recall and thus fails to utilize the unlabeled data to its full potential. Finding a good balance between high recall and high-quality pseudo labels is crucial for the success of semi-supervised 3D object detection. In addition, compared to their 2D counterparts, 3D datasets are substantially smaller in scale, mainly due to the increased effort required to collect the data. For instance, even the largest 3D indoor benchmark dataset, SUN RGB-D \cite{song2015sunrgbd}, contains only 10,335 samples. This limited scale significantly restricts the data diversity and hinders the overall model performance.

To this end, we present a novel approach for semi-supervised 3D object detection named Dual-Perspective Knowledge Enrichment (DPKE). DPKE aims to enrich the knowledge of limited training data, particularly unlabeled data, from two perspectives: data-perspective and feature-perspective. Specifically, for the data-perspective, we adopt data augmentation inspired by semi-sampling \cite{wu2022boosting} that randomly samples instances from labeled data and inserts them into the current training samples, to enhance the diversity of the input data, thereby enriching the knowledge of the data level. Unlike the uniform sampling for each class in \cite{wu2022boosting}, we design a class-probabilistic data augmentation method that selects sampled instances based on the varying distribution of class probabilities. These probabilities are computed from the model predictions, which are able to reflect the variance of class imbalance and learning difficulty. 
Our class-probabilistic data augmentation method can be viewed as a form of curriculum learning, as it gradually increases the sampling probability of instances in terms of classes with increasing difficulty during the training process as the model capacity improves.

Additionally, we introduce feature-perspective knowledge enrichment to compensate for  the high-quality yet low-recall pseudo labels as in 3DIoUMatch.
This is achieved by proposing a geometry-aware feature matching approach that can effectively exploit knowledge from teacher proposals with lower confidence. Instead of applying consistency regularization on model outputs like SESS, DPKE enforces consistency regularization on proposal features, as they contain richer information regarding potential objects. Before matching proposal features, a lower threshold is adopted in terms of objectness score to filter out background proposals. Subsequently, the geometry similarity between the points from the bounding box of the two proposals is measured, and this similarity is used to weigh the feature matching. 
This is because proposals with similar geometry are more likely to represent the same object, and therefore their features should be matched more strongly during the knowledge enrichment process.


The \textbf{main contribution} of this paper can be summarized as: 1) We propose a novel Dual-Perspective Knowledge Enrichment approach to address the challenges in semi-supervised 3D object detection, such as limited data diversity and unsatisfied pseudo labels. 2) We design a class-probabilistic data augmentation method for the data perspective and geometry-aware feature matching method for the feature perspective. 3) Our proposed method achieves state-of-the-art performance on two benchmark datasets for semi-supervised 3D object detection, demonstrating the effectiveness of our approach in enriching the knowledge from limited data.

\section{Related Work} \label{sec:related_work}

\paragraph{Semi-supervised 2D object detection.}
Many recent studies \cite{jeongConsistencybasedSemisupervisedLearning2019,chenDenseLearningBased2022,liuMixTeacherMiningPromising2023, liuUNBIASEDTEACHERSEMISUPERVISED2021, liuUnbiasedTeacherV22022, sohnSimpleSemiSupervisedLearning2020, tangHumbleTeachersTeach2021, xuEndtoEndSemiSupervisedObject2021, zhangSemiSupervisedObjectDetection2021, zhouDenseTeacherDense2022} have explored the realm of semi-supervised 2D object detection. Prior works have drawn insights from traditional semi-supervised learning techniques in the context of image classification. For instance, 
STAC \cite{sohnSimpleSemiSupervisedLearning2020} employs a two-stage training scheme for Faster-RCNN, where the first stage model provides pseudo labels for the subsequent stage. 
Subsequent works widely adopt the pseudo labeling strategy, leveraging the student-teacher paradigm \cite{xuEndtoEndSemiSupervisedObject2021,liuUNBIASEDTEACHERSEMISUPERVISED2021,liuUnbiasedTeacherV22022,li2022pseco,chen2022semi}, and focus on developing novel techniques to better utilize the teacher model's supervision signal. 
\cite{liuUNBIASEDTEACHERSEMISUPERVISED2021} presents the Unbiased Teacher and a class-balance loss to eliminate pseudo-labeling bias. Unbiased Teacher V2 \cite{liuUnbiasedTeacherV22022} further introduces the Listen2Student mechanism for anchor-free detectors to address the ineffectiveness of pseudo-labeling bounding box regression. PseCo \cite{li2022pseco} concentrates on merging pseudo labeling and consistency training, proposing a noisy pseudo box learning method for robust label assignment and localization quality evaluation. VCL \cite{chen2022semi} offers a solution that leverages confusing samples without label correction by assigning a virtual category to each sample.

\paragraph{Semi-supervised 3D object detection.}
Traditional semi-supervised learning techniques also benefit semi-supervised 3D object detection. However, due to disparities in data domain and model structure, state-of-the-art semi-supervised 2D object detection methods cannot be directly or trivially applied to the 3D domain. Current approaches \cite{zhao2020sess, wang20213dioumatch, wang2021temporal, park2022detmatch, zhang2022atf, yin2022proficient, chen2023class, wu2022boosting} mainly build upon frameworks introduced by earlier studies \cite{zhao2020sess,wang20213dioumatch}. 
One significant difference between semi-supervised detection in the 2D and 3D domains is the distinct nature of the backbones or techniques employed for indoor and outdoor scenes in 3D domain, due to the unique characteristics and complexities of indoor and outdoor environments. 
While semi-supervised 3D object detection in outdoor scenes \cite{wang20213dioumatch, wang2021temporal,park2022detmatch,zhang2022atf,yin2022proficient} has shown promising results, the performance of its indoor counterpart remains sub-optimal.
Several methods have been proposed to improve semi-supervised 3D object detection in indoor scenarios. Some methods, such as \cite{chen2023class, wang20213dioumatch}, focus on improving pseudo labeling to enhance performance. 3DIoUMatch \cite{wang20213dioumatch} designs an IoU branch and applies high thresholds to select high-quality pseudo labels, but at the cost of reduced pseudo label recall. 

\section{Methodology}\label{sec:method}


\subsection{Problem Definition}
In 3D semi-supervised detection, the goal is to leverage both labeled and unlabeled data for training detection models. The labeled data consists of $N_L$ samples, represented as $\{X_L^i, Y_L^i\}_{i=1}^{N_L}$, where $X_L^i$ represents $i$-th point cloud and $Y_L^i$ represents the corresponding annotation.  
The annotation $Y_L$ includes information such as the category label $y_{cls}$, the center location $c_x, c_y, c_z$, the size $s_x, s_y, s_z$, and the orientation $\theta$ of interested objects.
In addition to the labeled data, we also have $N_U$ unlabeled data, denoted as $\{X_U^i\}_{i=1}^{N_U}$. 
These unlabeled samples do not have corresponding annotations but can still be utilized to improve the performance of the detection models.
Note that $N_U$ and $N_L$ are determined by the label ratios set in the experiments, but typically follow the condition $N_U \gg N_L$.

\subsection{Dual-Perspective Knowledge Enrichment}
Our DPKE approach, as shown in Figure \ref{fig:pipeline}, follows the student-teacher paradigm \cite{tarvainen2017mean} commonly used in existing semi-supervised 3D detection methods \cite{wang20213dioumatch, wu2022boosting, zhao2020sess}. Both the student and teacher models share the same backbone architecture, specifically an improved version of VoteNet \cite{votenet} from \cite{wang20213dioumatch}. The student and teacher models are initialized with 
parameters exclusively trained  
on labeled data during the pre-training stage. In the training stage, the student model's weights are updated using the loss function gradients, while the teacher model's weights are updated using the exponential moving average of the student model's weights.

Following previous works \cite{wang20213dioumatch, wu2022boosting, zhao2020sess}, our DPKE utilizes different augmentation methods for the student and teacher models.
The teacher model undergoes weak augmentation of sub-sampling on the input point clouds. On the other hand, the student model is subjected to a strong augmentation scheme, including random flipping and scaling of the point clouds. However, unlike traditional scene level augmentation techniques, our DPKE introduces a class-probabilistic data augmentation method. This method adaptively augments instances into the scenes based on their class probabilities, leading to data-perspective knowledge enrichment. 

Our DPKE combines the high-precision yet low-recall pseudo labels used in 3DIoUMatch with knowledge from less confident proposals to improve overall performance. To achieve this, our DPKE introduces a geometry-aware feature matching method. 
This method applies adaptive consistency regularization on the features of potential foreground proposals, based on the geometry similarity between points within the proposal bounding boxes. The computed geometry-aware feature matching loss is combined with the original loss used in 3DIoUMatch to optimize the student model's parameters.

\subsection{Class-Probabilistic Data Augmentation}\label{sec:cpda}
Our class-probabilistic data augmentation method draws inspiration from semi-sampling \cite{wu2022boosting} but does not use the instance segmentation annotations in the ScanNet dataset. Instead, we employ a process where we randomly crop bounding boxes from labeled scenes, creating a \textit{proposal bank}. These bounding boxes are then inserted into the present scenes, similar to the gt-sampling \cite{yan2018second} technique used in fully-supervised 3D outdoor object detection. During the insertion process, we conduct collision detection to ensure that the inserted boxes do not intersect with existing objects in the scenes. However, since we do not have ground truth bounding boxes for the unlabeled scenes, we use the prediction results from the unlabeled scenes for collision detection. This allows us to leverage the information from the unlabeled data in a meaningful way during the data augmentation process.

Both semi-sampling and gt-sampling approaches typically apply data augmentation in a class-uniform manner. However, in the case of indoor scenes, where there are multiple categories with different learning statuses \cite{chen2023class}, it becomes necessary to assign different sampling probabilities to each category. By incorporating category-specific sampling probabilities, our class-probabilistic data augmentation method can better balance the augmentation process and address the specific learning needs of each category in the indoor scene dataset. Specifically, to assign different sampling probabilities to each category, we utilize the average logits for each category during training. These logits are then normalized into probabilistic weights, ranging from 0 to 1. We achieve this normalization by first applying min-max normalization to ensure the values are within the desired range and subsequently applying the sigmoid function to obtain the final probabilistic weights. 
We do not use the softmax function to obtain the probabilistic weights because its exponential operation would amplify differences between input values and produce a sharper/distorted output, which harms the diversity of the augmented scenes.  
Subsequently, we use the obtained probabilities to sample the proposals from the proposal bank and paste it into the existing scene at the first $N$ epochs. The experimental results in Figure~\ref{fig:diff} show the superiority of our class-probabilistic data augmentation method over the class-uniform semi-sampling.

\paragraph{Additional Analysis.}
We analyze our class-probabilistic data augmentation method by concentrating on the seed features generated during the intermediate stage of VoteNet \cite{votenet}. A detailed description of seed features can be found in the Appendix A. Formally, $P(y|f_c, f_n)$ represents the predicted probabilities of various predictions $y$, such as the locations and classes of proposals, given the seed features within the target proposals ($f_c$) and the seed features outside the target proposals ($f_n$). Here we expect the model to learn the relations between $f_c$ and $y$, which is meaningful. By leveraging the relations between ($f_c$, $f_n$) and $y$, we can derive the following equation:
\begin{equation}
    P(y|f_c, f_n ) = P(f_c, y) \cdot \frac{P(f_n|y,f_c)}{P(f_n, f_c)}.
\end{equation}
$P(f_c, y)$ represents the joint probability of $f_c$ and the prediction $y$, which the model aims to learn. However, considering that $f_n$ is unrelated to $f_c$ and $y$, the term $P(f_n|y,f_c)/P(f_n, f_c)$ captures the potential for significant fluctuations when the model struggles to effectively distinguish between $f_c$ and $f_n$. These fluctuations can impact the training process as it now must accommodate the additional variability introduced by the $f_n$ features. 
To address this issue, our sampling strategy that pastes more well-learned object proposals into existing scenes appears to be a feasible solution, which can simultaneously balance enhancing diversity and improving model learning stability. 
%
%

\subsection{Geometry-aware Feature Matching}\label{sec:gafm}
3DIoUMatch \cite{wang20213dioumatch} adopts high thresholds to select pseudo labels with high precision. However, it comes at the cost of decreasing the recall of pseudo labels. To address this, we propose a new geometry-aware feature matching method that focuses on enriching knowledge from proposals with lower confidence. Our method introduces geometry-aware weights for each proposal alignment to regulate the discrepancy in proposal features between the student and teacher models. 

The presence of random down-sampling steps, such as farthest point sampling (FPS), in the model backbone leads to different sets of proposals for the student and teacher models when given the same input point cloud. Consequently, the alignment of proposals between the two models cannot be guaranteed. To address this issue, we follow \cite{zhao2022static} to save the indexes of selected points from the student model, which are reused to sample points in FPS when processing the same input with the teacher model. This ensures that the predicted results of the teacher model and student model are aligned, enabling the computation of a feature matching loss.

\begin{table*}[t]
\resizebox{\textwidth}{!}{
\begin{tabular}{|c|c|cc|cc|cc|cc|}
\hline
\multirow{2}{*}{Dataset}   & \multirow{2}{*}{Model}                & \multicolumn{2}{c|}{5\%}                             & \multicolumn{2}{c|}{10\%}                            & \multicolumn{2}{c|}{20\%}                            & \multicolumn{2}{c|}{100\%}              \\ \cline{3-10} 
                           &                                       & \multicolumn{1}{c|}{mAP@0.25}       & mAP@0.5        & \multicolumn{1}{c|}{mAP@0.25}       & mAP@0.5        & \multicolumn{1}{c|}{mAP@0.25}       & mAP@0.5        & \multicolumn{1}{c|}{mAP@0.25} & mAP@0.5 \\ \hline \hline
\multirow{7}{*}{ScanNet}   & VoteNet \cite{votenet}                & \multicolumn{1}{c|}{27.9}           & 10.8           & \multicolumn{1}{c|}{36.9}           & 18.2           & \multicolumn{1}{c|}{46.9}           & 27.5           & \multicolumn{1}{c|}{57.8}     & 36.0    \\ \cline{2-10} 
                           & SESS \cite{zhao2020sess}              & \multicolumn{1}{c|}{--}             & --             & \multicolumn{1}{c|}{39.7}           & 18.6           & \multicolumn{1}{c|}{47.9}           & 26.9           & \multicolumn{1}{c|}{62.1}     & 38.8    \\ \cline{2-10} 
                           & 3DIoUMatch* \cite{wang20213dioumatch} & \multicolumn{1}{c|}{39.44$\pm$1.15} & 22.19$\pm$1.11 & \multicolumn{1}{c|}{47.53$\pm$1.78} & 28.63$\pm$1.41 & \multicolumn{1}{c|}{52.70$\pm$1.56} & 35.33$\pm$0.7  & \multicolumn{1}{c|}{62.81}    & 42.16   \\ \cline{2-10} 
                           & Confid-3DIoUMatch* \cite{chen2023class}    & \multicolumn{1}{c|}{39.89$\pm$0.63} & 23.96$\pm$0.36 & \multicolumn{1}{c|}{47.90$\pm$2.73} & 29.29$\pm$1.45 & \multicolumn{1}{c|}{53.17$\pm$0.42} & 35.84$\pm$0.31 & \multicolumn{1}{c|}{61.70}    & 41.67   \\ \cline{2-10} 
                           & Semi-Sampling* \cite{wu2022boosting}  & \multicolumn{1}{c|}{41.94$\pm$2.32} & 24.86$\pm$1.52 & \multicolumn{1}{c|}{48.32$\pm$0.85} & 30.15$\pm$0.76 & \multicolumn{1}{c|}{55.39$\pm$1.06} & 37.61$\pm$2.01 & \multicolumn{1}{c|}{64.49}    & 47.46   \\ \cline{2-10} 
                           & \textbf{Ours}                         & \multicolumn{1}{c|}{\textbf{44.01$\pm$1.07}} & \textbf{27.04$\pm$1.88} & \multicolumn{1}{c|}{\textbf{51.88$\pm$1.40}} & \textbf{34.06$\pm$0.71} & \multicolumn{1}{c|}{\textbf{57.64$\pm$0.80}} & \textbf{41.35$\pm$1.07} & \multicolumn{1}{c|}{\textbf{65.33}}    & \textbf{48.71}   \\ \cline{2-10} 
                           & improvements                          & \multicolumn{1}{c|}{2.07}           & 2.18           & \multicolumn{1}{c|}{3.56}           & 3.91           & \multicolumn{1}{c|}{2.25}           & 3.74           & \multicolumn{1}{c|}{0.84}     & 1.25    \\ \hline\hline
\multirow{7}{*}{SUN RGB-D} & VoteNet \cite{votenet}                & \multicolumn{1}{c|}{29.9}           & 10.5           & \multicolumn{1}{c|}{38.9}           & 17.2           & \multicolumn{1}{c|}{45.7}           & 22.5           & \multicolumn{1}{c|}{58.0}     & 33.4    \\ \cline{2-10} 
                           & SESS \cite{zhao2020sess}              & \multicolumn{1}{c|}{--}             & --             & \multicolumn{1}{c|}{42.9}           & 14.4           & \multicolumn{1}{c|}{47.9}           & 20.6           & \multicolumn{1}{c|}{61.1}     & 37.3    \\ \cline{2-10} 
                           & 3DIoUMatch* \cite{wang20213dioumatch} & \multicolumn{1}{c|}{38.72$\pm$1.20} & 21.31$\pm$1.67 & \multicolumn{1}{c|}{46.02$\pm$0.53} & 28.88$\pm$0.59 & \multicolumn{1}{c|}{50.39$\pm$0.85} & 30.71$\pm$0.48 & \multicolumn{1}{c|}{61.77}    & 41.85   \\ \cline{2-10} 
                           & Confid-3DIoUMatch* \cite{chen2023class}    & \multicolumn{1}{c|}{38.95$\pm$1.71} & 21.81$\pm$0.76 & \multicolumn{1}{c|}{45.92$\pm$0.82} & 28.91$\pm$0.32 & \multicolumn{1}{c|}{50.43$\pm$1.46} & 30.57$\pm$1.14 & \multicolumn{1}{c|}{60.14}    & 40.38   \\ \cline{2-10} 
                           & Semi-Sampling* \cite{wu2022boosting}  & \multicolumn{1}{c|}{39.79$\pm$1.08} & 22.60$\pm$0.55 & \multicolumn{1}{c|}{48.57$\pm$0.69} & 31.06$\pm$0.49 & \multicolumn{1}{c|}{51.43$\pm$1.15} & 33.21$\pm$0.39 & \multicolumn{1}{c|}{63.24}    & 45.73   \\ \cline{2-10} 
                           & \textbf{Ours}                         & \multicolumn{1}{c|}{\textbf{41.51$\pm$0.99}} & \textbf{24.98$\pm$1.2}  & \multicolumn{1}{c|}{\textbf{49.93$\pm$0.98}} & \textbf{32.48$\pm$0.4}  & \multicolumn{1}{c|}{\textbf{53.26$\pm$0.19}} & \textbf{35.01$\pm$0.22} & \multicolumn{1}{c|}{\textbf{63.92}}    & \textbf{46.87}   \\ \cline{2-10} 
                           & improvements                          & \multicolumn{1}{c|}{1.72}           & 2.38           & \multicolumn{1}{c|}{1.36}           & 1.42           & \multicolumn{1}{c|}{1.83}           & 1.80           & \multicolumn{1}{c|}{0.68}     & 1.14    \\ \cline{1-10} 
\end{tabular}
}
\caption{\small{\textbf{Comparison with previous methods on ScanNet and SUN RGB-D datasets}. Mark * indicates that the result is reproduced by ourselves under the same data splits with 3DIoUMatch's.}}
\label{tab:main}
\end{table*}
\begin{table*}[]
    \centering
    \resizebox{\textwidth}{!}{
        \begin{tabular}{c|cc|cc|cc|cc}
\hline \hline
\multirow{2}{*}{ID} & \multicolumn{2}{c|}{Class-Prob. Data Aug.} & \multicolumn{2}{c|}{Geometriy-aware Feat. Matching} & \multicolumn{2}{c|}{ScanNet 10\%} & \multicolumn{2}{c}{SUN-RGBD 5\%} \\ \cline{2-9} 
& uniform sampling                  & class-prob. strategy                    & feature matching            & geo. weights              & mAP@0.25         & mAP@0.5        & mAP@0.25        & mAP@0.5        \\ \hline \hline
(a)                 &                              &                             &                           &                           & 47.53            & 28.63          & 38.72           & 21.31          \\ \hline
(b)                 & \checkmark    &                             &                           &                           & 48.32            & 30.15          & 39.79           & 22.60          \\ \hline
(c)                 & \checkmark    & \checkmark   &                           &                           & 50.63            & 31.72          & 40.35           & 23.75          \\ \hline
(d)                 & \checkmark    &                             & \checkmark & \checkmark & 50.82            & 32.60          & 40.79           & 23.90          \\ \hline
(e)                 & \checkmark    & \checkmark   & \checkmark & \checkmark & \textbf{51.88}            & \textbf{34.06}          & \textbf{41.51}           & \textbf{24.98}          \\ \hline \hline
(f)                 &                              &                             &                           &                           & 47.53            & 28.63          & 38.72           & 21.31          \\ \hline
(g)                 &                              &                             & \checkmark &                           & 48.59            & 29.34          & 39.19           & 21.86          \\ \hline
(h)                 &                              &                             & \checkmark & \checkmark & 49.68            & 31.16          & 39.44           & 22.29          \\ \hline
(i)                 & \checkmark    & \checkmark   & \checkmark &                           & 50.89            & 32.37          & 40.79           & 24.27          \\ \hline
(j)                 & \checkmark    & \checkmark   & \checkmark & \checkmark & \textbf{51.88}            & \textbf{34.06}          & \textbf{41.51}           & \textbf{24.98}          \\ \hline \hline
\end{tabular}
    }
    \caption{\small{\textbf{Ablation Study on ScanNet and SUN-RGBD datasets.} Comprehensive ablation experiments are conducted on the proposed DPKE, validating the independent effectiveness of each individual module.}}
    \label{tab:ablation}
\end{table*}

After aligning the proposals, we obtain the proposal features $z^s$ and $z^t$ for the student and teacher models, respectively. 
These features are obtained by concatenating the features from all three convolutional layers in the proposal module \cite{votenet}. The feature matching loss is computed as:
\begin{equation}
    L_{\delta}(z^s, z^t) = \left\{
    \begin{aligned}
         & \frac{1}{2}(z^s - z^t)^2,                &  & \text{if } |z^s - z^t| \leq \delta, \\
         & \delta|z^s - z^t| - \frac{1}{2}\delta^2, &  & \text{otherwise.}
    \end{aligned}
    \right.
\end{equation}
To focus on informative proposals, we apply an objectness score threshold $\tau_{\text{obj}}$ that is lower than the one used for selecting high-precision pseudo labels. This allows us to exclude numerous background proposals that may not contribute significantly to knowledge enrichment. Additionally, in complex scenes where objects can overlap, even slight changes in the locations and sizes of proposals can result in noticeable variations in geometry properties. We believe that proposal pairs with similar geometry properties should undergo stronger regularization than those with dissimilar geometry. To incorporate the similarity of geometry properties, we utilize the Chamfer distance \cite{fan2017point} to calculate the similarity between point sets within the two proposals. Let $P^t \in \mathbb{R}^{n_t \times 3}$ and $P^s \in \mathbb{R}^{n_s \times 3}$ represent the point sets in the predicted proposals of the teacher model and student model, respectively, with $n_t$ and $n_s$ denoting the number of points in each set. Based on the Chamfer distance, we can calculate the geometric-aware weight for a proposal pair as follows:
%
%
\begin{equation}
\begin{aligned}
    \mathcal{W}_{\text{geometry}} = \exp(-\sum\limits_{i=1}^{n_t} & \min\limits_{P^s_j\in P^s} \lVert P^t_i - P^s_j \rVert_2^2 - \\
    & \sum\limits_{i=1}^{n_s} \min\limits_{P^t_j\in P^t} \lVert P^s_i - P^t_j \rVert_2^2).
\end{aligned} 
\end{equation}
However, directly calculating $\mathcal{W}_{\text{geometry}}$ may consume too many computational resources because we have no limitations on the number of points in $P^t$ and $P^s$. 
We address this challenge by developing an efficient computation method, which is detailed in the Appendix B. 
Finally, the geometry-aware feature matching loss can be computed as:
\begin{equation}
    \mathcal{L}_f = \mathbb{1}(o^t \geq \tau_{\text{obj}}) \cdot \mathcal{L}_\delta(z^s, z^t) \cdot \mathcal{W}_{\text{geometry}},
\end{equation}
where $o^t$ is the objectness score predicted by the student model, $\mathbb{1}(x)$ takes a value of 1 only when the condition $x$ is true; otherwise, it is 0.
%

\section{Experiments}\label{sec:exp}

\begin{table*}[t]
    \centering
    \begin{subtable}[t]{.45\textwidth}
    \centering
    \resizebox{\textwidth}{!}{
    \begin{tabular}{lcccc}
    \toprule
    \multirow{2}{*}{strategy}& \multicolumn{2}{c}{ScanNet 10\%} & \multicolumn{2}{c}{SUN-RGBD 5 \%} \\ 
     &   \fontsize{8}{8}\selectfont mAP@0.25      & \fontsize{8}{8}\selectfont mAP@0.5        & \fontsize{8}{8}\selectfont mAP@0.25         & \fontsize{8}{8}\selectfont mAP@0.5        \\ \midrule
    Uniform                    & 48.32           & 30.15          & 39.79            & 22.60          \\
    LLS                      &  48.07          &  29.32         &   39.42       &   22.39        \\
    HLS (\textbf{ours})                     & \textbf{50.63}           & \textbf{31.72}          & \textbf{40.35}            & \textbf{23.75}          \\ \bottomrule
    \end{tabular}
    }
    \caption{\small{Different sampling strategies for Class-Probabilistic Data Augmentation. Here LLS is short for Low-Logit Sampling and HLS is short for High-Logit Sampling.}}
    \end{subtable}
    ~
    \begin{subtable}[t]{.45\textwidth}
    \centering
    \resizebox{\textwidth}{!}{
    \begin{tabular}{lcccc}
    \toprule
    \multirow{2}{*}{weighting}& \multicolumn{2}{c}{ScanNet 10\%} & \multicolumn{2}{c}{SUN-RGBD 5 \%} \\ 
     &   \fontsize{8}{8}\selectfont mAP@0.25      & \fontsize{8}{8}\selectfont mAP@0.5        & \fontsize{8}{8}\selectfont mAP@0.25         & \fontsize{8}{8}\selectfont mAP@0.5        \\ \midrule
    Constant                    & 48.59           & 29.34          & 39.19            & 21.86          \\
    HCD                      &    48.83        &  30.19         &    39.31         &   21.99        \\
    LCD (\textbf{ours})                      & \textbf{49.68}           & \textbf{31.16}          & \textbf{39.44}            & \textbf{22.29}          \\ \bottomrule
    \end{tabular}
    }
    \caption{\small{Different weighting strategies for geometry-aware feature matching. LCD means giving samples with Low Chamfer Distance higher weights. HCD is short for High Chamfer Distance, thus reverse.}}
    \end{subtable}
\caption{\small{Further validation of the effectiveness of DPKE.}}
\label{tab:further_ablation}
\end{table*}

\begin{figure*}[t]
    \centering
    \includegraphics[width=.85\textwidth]{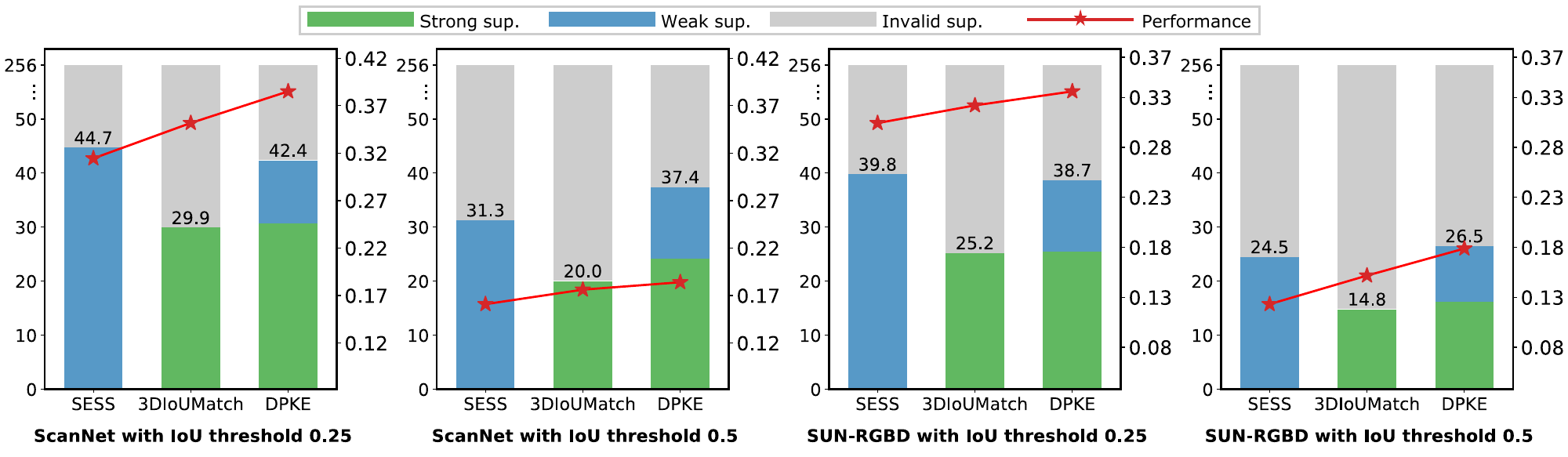}
    \caption{\small{The distribution of samples providing different supervision from the teacher model predictions. The invalid supervision represents the samples failing to match ground truth. The rest part could be divided into samples providing strong (e.g. 3DIoUMatch) and weak supervision (e.g. SESS or our Geometry-aware Feature Matching).}}
    \label{fig:convergence}
\end{figure*}
\begin{figure*}[t]
    \centering
    \includegraphics[width=.85\textwidth]{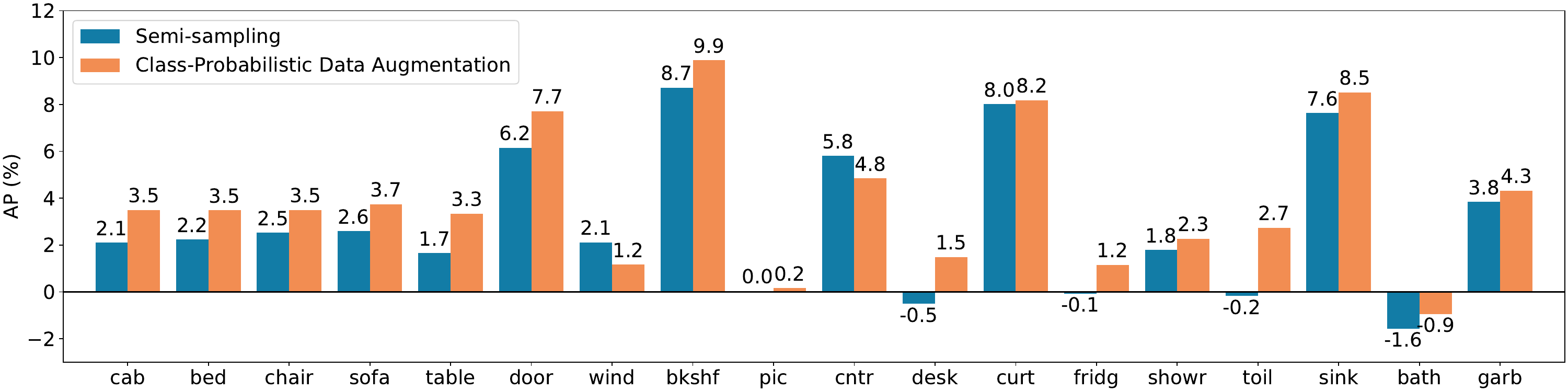}
    \caption{\small{Per class average precision (AP) comparison by applying different data augmentation strategies on 3DIoUMatch. The experiment is conducted on the ScanNet with the label ratio as 5\%.}}
    \label{fig:diff}
\end{figure*}


\subsection{Evaluation Setup}\label{sec:implementation}

\paragraph{Datasets and Training Details.} We validate our proposed method on the ScanNet \cite{dai2017scannet} and SUN RGB-D \cite{song2015sunrgbd} datasets.
Our code is built based on \cite{wang20213dioumatch}, adhering to their optimization and training parameters. We conduct experiments with label ratios of 0.05, 0.1, 0.2, and 1.0, as presented in Table~\ref{tab:main}. For both the ScanNet and SUN RGB-D datasets, we follow the standard preprocessing \cite{zhao2020sess, wang20213dioumatch} and evaluation protocols to ensure consistency across different experiments. These protocols involve reporting the mean average precision (mAP) over three data splits with 3D Intersection over Union (IoU) thresholds of 0.25 and 0.5. Further details on implementation and datasets can be found in the Appendix B.

\paragraph{Methods.} We compare our method with five other methods, as displayed in Table~\ref{tab:main}. Both 3DIoUMatch~\cite{wang20213dioumatch} and Confid-3DIoUMatch~\cite{chen2023class} have open-source code. We do experiments using their code on the same splits and report their performance. Semi-sampling \cite{wu2022boosting} does not have a codebase available online. We contact their authors and reproduce their results, removing the extra segmentation annotations on ScanNet.

\begin{figure*}[t]
    \centering
    \includegraphics[width=.74\textwidth]{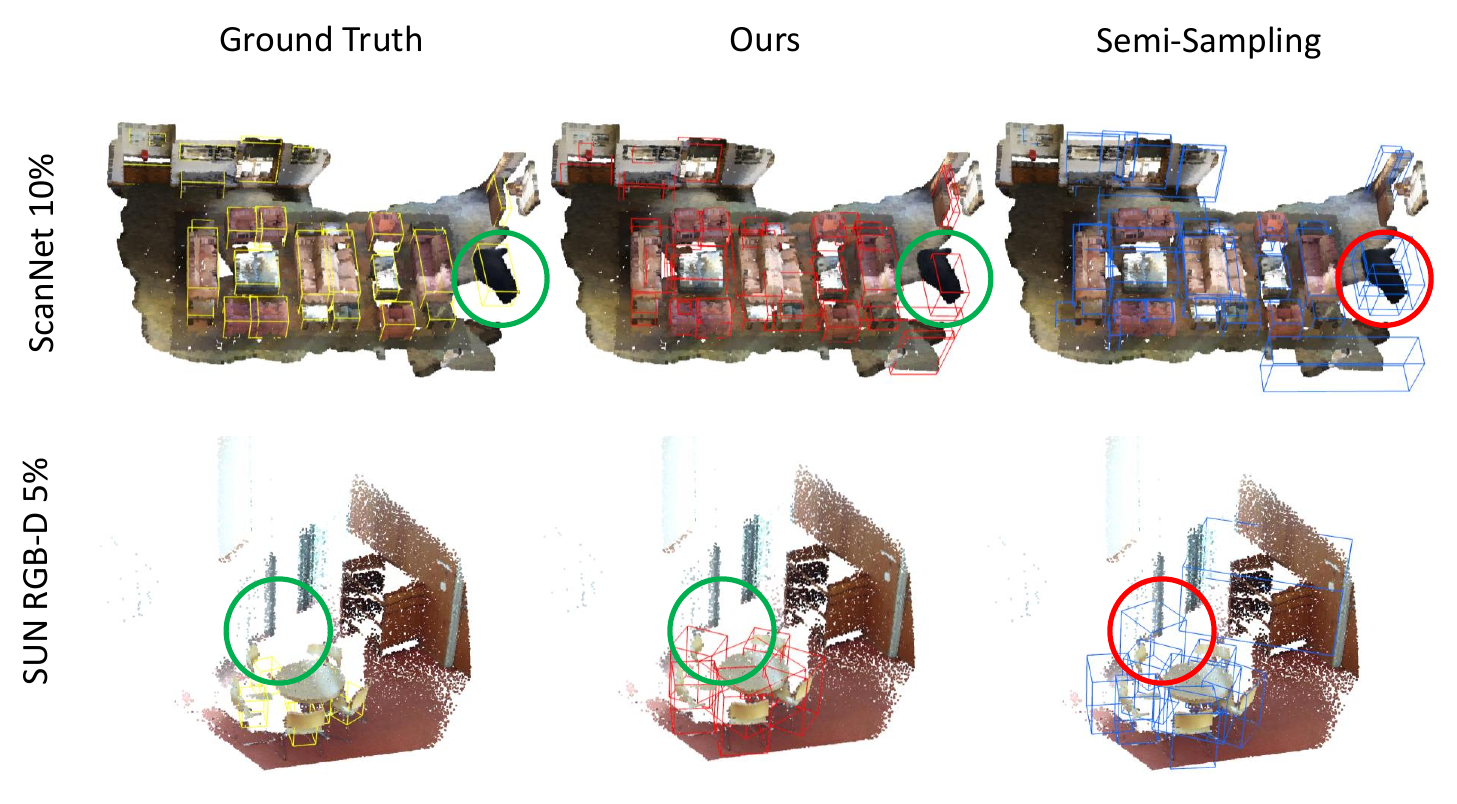}
    \caption{\small{\textbf{Visualization results on ScanNet and SUN-RGBD.} The models are trained on ScanNet 5\% and SUN RGB-D 10\%, respectively. The areas within the green circles contain the ground truth and correct predictions, while those in the red circles contains wrong predictions.}}
    \label{fig:visualization}
\end{figure*}

\subsection{Main Results}
Table~\ref{tab:main} presents the comparison results against baselines under different ratios of labeled data. The previous state-of-the-art is semi-sampling~\cite{wu2022boosting}, which verifies that the diversity of input data is the most severe problem for semi-supervised 3D detection. The performance of Confid-3DIoUMatch~\cite{chen2023class} is much weaker than semi-sampling. This is likely because the resampling and reweighting strategies in Confid-3DIoUMatch are still scene-level, which cannot increase the diversity of input data compared to semi-sampling. Compared with these methods, our method achieves 2\%-4\% increase on ScanNet. Even on the much more difficult dataset SUN RGB-D~\cite{song2015sunrgbd}, we still achieve around 2\% increase under all settings on SUN RGB-D. 
We hypothesize that the weaker improvement observed on the SUN RGB-D dataset, in comparison to the ScanNet dataset, is attributable to the inferior quality of point clouds and ground truth proposals. Our class-probabilistic data augmentation is impacted on SUN RGB-D, as it depends on ground truth information to generate the proposal bank. Additionally, the geometry-aware feature matching loss is also influenced by the considerable incompleteness present in low-quality point clouds. 
Notably, under the fully labeled ratio, our method achieves only 1\% increase for both ScanNet and SUN RGB-D, as the geometry-aware feature matching loss remains inactive in the fully labeled data setting.

\subsection{Ablation Study}
\paragraph{Ablation of class-probabilistic data augmentation.}
To validate the effectiveness of our proposed class-probabilistic data augmentation, we test a degraded version of our proposed algorithm, which uses uniform sampling. Then compare it to the results obtained when applying our class-probabilistic augmentation strategy. The experiments are done on ScanNet 10\% and SUN RGB-D 5\%, as displayed in Table~\ref{tab:ablation} from (a) to (e). (a) shows the performance of the 3DIoUMatch. The results with or without geometry-aware feature matching loss are shown in (b) and (c), or (d) and (e). 
We also compare our class-probabilistic sampling strategy with different sampling strategies in Table~\ref{tab:further_ablation} (a). The performance indicates that forcing the model to learn from examples with low logits even harms the mAP under both ScanNet and SUN RGB-D. 
That is because the categories with lower logits are usually the ones that are difficult to recognize, such as ``picture''. 
Utilizing such categories to augment the current scene does not effectively serve the purpose of increasing diversity.

\paragraph{Ablation of geometry-aware feature matching.} Our proposed geometry-aware feature matching loss consists of the feature matching loss and geometry weights for each proposal pair. We present ablation studies for each component under various conditions in Table~\ref{tab:ablation}, ranging from (f) to (j).
Regardless of whether class-probabilistic data augmentation is applied, each module in our geometry-aware feature matching loss can contribute to improvements independently.
Our geometry-aware constraints encourage the model to focus more on proposal pairs with lower Chamfer Distance, denoted as LCD. To validate our method's efficacy, we compare it to an alternative weighting strategy that assigns higher weight to proposal pairs with higher Chamfer Distance. The ``constant'' means the weights for all proposal pairs equals to one. As demonstrated in Table~\ref{tab:further_ablation} (b), our method outperforms HCD and constant weights under all settings. 

\paragraph{Pseudo-label convergence.}
Our DPKE can effectively utilize more supervision signals. As illustrated in Figure~\ref{fig:convergence}, we collect statistics on the teacher model's output during the training process, focusing on supervision signals from samples that can be matched with the ground truth, namely those corresponding to the blue and green portions. The other samples are represented by the gray protion and cannot provide valid supervision to the student model. The red line illustrates the performance of SESS, 3DIoUMatch, and DPKE, respectively. Under various settings, our method is able to utilize more valid samples predicted by the teacher model compared to 3DIoUMatch, and is more efficient than SESS.

\paragraph{Class-wise Performance Comparison with Different Augmentations.}
Figure~\ref{fig:diff} compares the mAP improvement for each class. Our method outperforms semi-sampling in all categories except "counter" and "window". This might be due to the fact that these two classes face the challenges of limited samples or high learning difficulty. 
For instance, the ScanNet training dataset contains only 216 ``counter" instances, which is significantly less than the ``chair" class containing 4,357 instances. On the other hand, the AP for ``window" is merely 12.4\% due to the learning difficulty. Our method assigns much lower probabilities to such categories, thus avoiding negative influences on other categories.

\subsection{Visualization}
Figure~\ref{fig:visualization} visualizes the qualitative comparison between our proposed method and the existing SOTA semi-sampling \cite{wu2022boosting} on a testing scene of ScanNet and SUN RGB-D, respectively. We observe that semi-sampling exhibits a tendency to predict redundant proposals, while our proposed DPKE aligns with ground truth well. For ScanNet, the semi-sampling predicts redundant proposals around the object in the circle. In SUN RGB-D, semi-sampling even predicts proposals in areas without distinct objects. This may be attributed to the lack of geometry-aware feature matching, leading to the model's inadequate perception of the geometry properties of objects.

\section{Conclusion}
In this study, we present DPKE, an approach that addresses the challenges in 3D semi-supervised object detection from both data and feature perspectives. We develop class-probabilistic data augmentation, which effectively handles the limited diversity in 3D datasets at the data level and minimizes potential inter-class influences resulting from varying learning progress. Furthermore, we introduce the geometry-aware feature matching loss to overcome the low recall of pseudo labels encountered in previous semi-supervised 3D detection techniques. Our geometry-aware feature matching resolves the low recall issue by utilizing feature-level similarity between student and teacher models, thereby enhancing the quantity of supervision. Comprehensive experiments on two benchmark datasets, ScanNet and SUN-RGBD, demonstrate that our proposed DPKE method surpasses existing state-of-the-art techniques under various label ratio conditions, highlighting its capacity to reduce the labor-intensive process of data annotation in cluttered 3D indoor environments.

\section{Acknowledgments}
The authors would like to thank the reviewers for their
comments that help improve the manuscript. This research work is supported by the National Research Foundation, Singapore under its AI Singapore Programme (AISG Award No: AISG2-RP-2021-022), the Hyundai research grant (04OIS000257C130), and the Agency for Science, Technology and Research (A*STAR) under its MTC Programmatic Funds (Grant No. M23L7b0021).


\bibliography{aaai24}

\let\titleold\title
\renewcommand{\title}[1]{\titleold{#1}\newcommand{\thetitle}{#1}}
\def\maketitlesupplementary
   {
   \newpage
       \twocolumn[
        \centering
        \Large
        \vspace{2.0em}Appendix \\
        \vspace{2.0em}
       ] 
   }
\maketitlesupplementary
\begin{appendix}

\begin{figure*}[t]
    \centering
    \includegraphics[width=.9\textwidth]{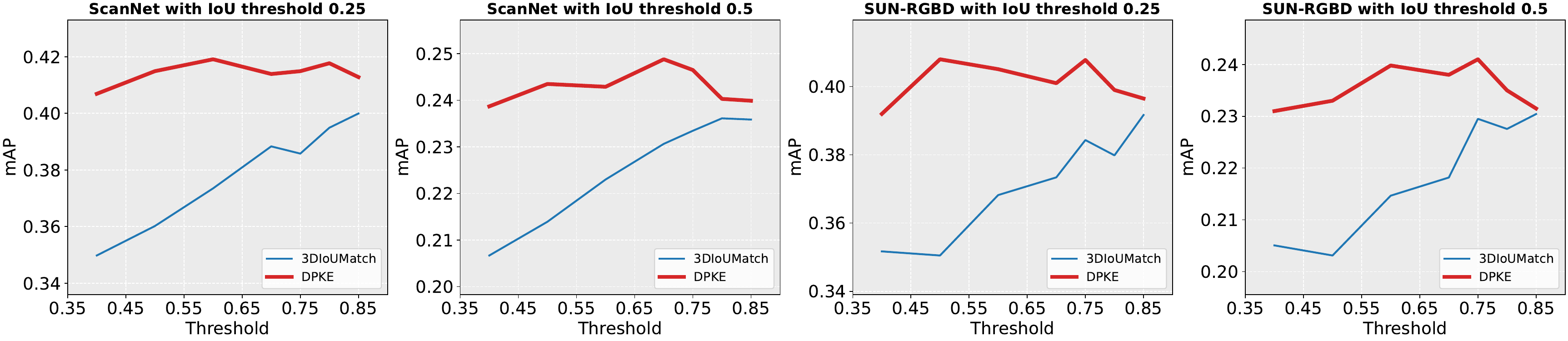}
    \caption{Performance of our proposed DPKE and 3DIoUMatch under different objectness thresholds.}
    \label{fig:threshold}
\end{figure*}
\begin{figure*}[t]
    \centering
    \includegraphics[width=.9\textwidth]{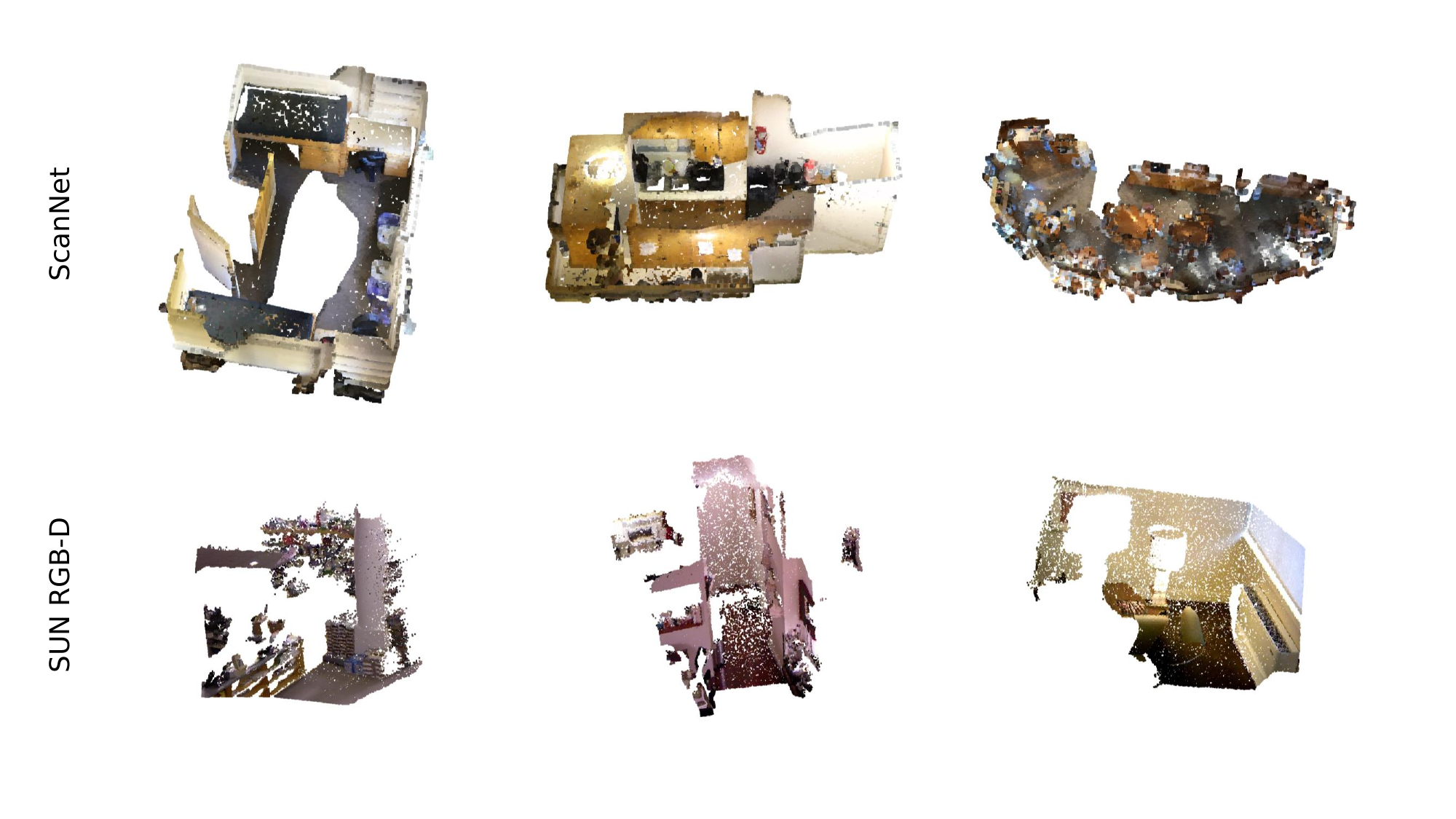}
    \caption{Some scenes of ScanNet and SUN RGB-D.}
    \label{fig:appendix}
\end{figure*}
\begin{table*}[t]
    \resizebox{\textwidth}{!}{%
        \begin{tabular}{c|c|cc|cc|cc}
            \hline
            \multirow{2}{*}{Dataset}   & \multirow{2}{*}{Model} & \multicolumn{2}{c|}{5\%} & \multicolumn{2}{c|}{10\%} & \multicolumn{2}{c}{20\%}                                                                               \\ \cline{3-8}
                                       &                        & mAP@0.25                 & mAP@0.5                   & mAP@0.25                 & mAP@0.5                 & mAP@0.25                & mAP@0.5                 \\ \hline\hline
            \multirow{3}{*}{ScanNet}   & 3DIoUMatch             & 39.44$\pm$1.15           & 22.19$\pm$1.11            & 47.53$\pm$1.78           & 28.63$\pm$1.41          & 52.70$\pm$1.56          & 35.33$\pm$0.7           \\
                                       & SESS+3DIoUMatch        & 38.85$\pm$1.31           & 21.77$\pm$2.45            & 47.03$\pm$1.12           & 27.81$\pm$1.37          & 52.12$\pm$0.5           & 33.15$\pm$0.70          \\
                                       & Ours                   & \textbf{44.01$\pm$1.07}  & \textbf{27.04$\pm$1.88}   & \textbf{51.88$\pm$1.40}  & \textbf{34.06$\pm$0.71} & \textbf{57.64$\pm$0.80} & \textbf{41.35$\pm$1.07} \\ \hline
            \multirow{3}{*}{SUN RGB-D} & 3DIoUMatch             & 38.72$\pm$1.20           & 21.31$\pm$1.67            & 46.02$\pm$0.53           & 28.88$\pm$0.59          & 50.39$\pm$0.85          & 30.71$\pm$0.48          \\
                                       & SESS+3DIoUMatch        & 37.04$\pm$1.46           & 20.56$\pm$1.79            & 45.77$\pm$1.93           & 28.13$\pm$0.69          & 49.26$\pm$1.21          & 29.11$\pm$0.36          \\
                                       & Ours                   & \textbf{41.51$\pm$0.99}  & \textbf{24.98$\pm$1.2}    & \textbf{49.93$\pm$0.98}  & \textbf{32.48$\pm$0.4}  & \textbf{53.26$\pm$0.19} & \textbf{35.01$\pm$0.22} \\ \cline{1-8}
        \end{tabular}
    }
    \caption{Comparison between our DPKE with the loss combining SESS's consistency loss and 3DIoUMatch's loss.}
    \label{tab:comparison}
\end{table*}

This appendix is organized as follows:

\begin{itemize}
    \item Section \ref{sec:detection_backbone} provides the detailed description of the backbone model \cite{votenet}.
    \item Section \ref{sec:dataset_and_implementation} provides additional details about the dataset and implementation.
    \item Section \ref{sec:additional} provides two additional experiments to verify the effectiveness of DPKE. The first one is about the sensativeness of objectness threshold. The second experiment is comparing our proposed DPKE with the loss combining SESS and 3DIoUMatch.
\end{itemize}

\section{Detection backbone}\label{sec:detection_backbone}

Building upon the foundational work \cite{wang20213dioumatch, chen2023class, wu2022boosting}, we employ VoteNet \cite{votenet} with an IoU branch as our detection framework. This structure is rooted in the PointNet++ architecture \cite{pointnet++}. Initially, VoteNet operates on an input point cloud with $N$ points sampled via random sampling, generating a rich sub-sample of seed point features. Following this, these seed points cast votes for their respective object centers, leading to the formation of $K$ clusters. These clusters are subsequently integrated to predict parameters for 3D bounding boxes, objectness scores, and probability distributions across various semantic classes. Throughout this procedure, the model performs sub-sampling of the point cloud features from $N$ to $K$ using Farthest Point Sampling (FPS). This ensures that even if the same input point clouds are processed twice, the output proposals might vary due to the inherent randomness of FPS. The bounding box parameters encompass the center location $c \in R3$, scale $d \in R3$, and orientation $\theta$ about the upright axis.
During its training phase, VoteNet simultaneously minimizes multiple target losses, including vote coordinate regression, objectness score binary classification, box center regression, bin classification, residual regression for heading angle, scale regression, and category classification. In the testing phase, VoteNet employs Non-Maximum Suppression (NMS) based on objectness scores to remove duplicate bounding boxes. For the 3DIoUMatch approach \cite{wang20213dioumatch}, a 3D IoU estimation module designed specifically for VoteNet is used. The IoU estimation branch does not feed gradients back to the feature backbone and is solely utilized for filtering metrics. We follow the model of 3DIoUMatch for fair comparison with previous works \cite{wang20213dioumatch,chen2023class,wu2022boosting}.
\section{Dataset and Implementation details}\label{sec:dataset_and_implementation}
\subsection{Dataset}
ScanNet and SUN RGB-D are leading indoor benchmark datasets widely utilized for 3D object detection and semantic segmentation tasks. While SUN RGB-D is larger in scale than ScanNet, the quality of each scene in SUN RGB-D is inferior to that in ScanNet, as illustrated in Figure~\ref{fig:appendix}. ScanNet scenes typically present more comprehensive and clearer views compared to SUN RGB-D, containing more complex structures, which enable a more comprehensive evaluation of the model's capabilities.

\paragraph{ScanNet.} ScanNet is an extensive indoor scene dataset comprising 1,513 reconstructed meshes from 707 unique indoor scenes. The dataset officially splits these scenes into 1,201 training samples and 312 validation samples. Each scene in the dataset is meticulously annotated with semantic segmentation masks, supplying detailed information for researchers to use in their experiments. The dataset includes 18 object classes out of the available 21 semantic classes. To generate the input point clouds, vertices are sampled from the meshes. As the ScanNet dataset does not include any existing amodal or orientated 3D bounding boxes, axis-aligned bounding boxes are derived from point-level labeling.

\paragraph{SUN RGB-D.} The SUN RGB-D dataset serves as an indoor benchmark for 3D object detection, comprising 10,335 single-view RGB-D images. These images are officially divided into 5,285 training samples and 5,050 validation samples. The dataset provides 3D bounding box annotations for multiple object classes, enabling a broad range of evaluation and comparison opportunities. Evaluation is performed on the 10 most common categories to facilitate comparisons with prior methods and models. By utilizing the camera parameters provided in the dataset, depth images are converted into point clouds, which serve as input for various models and algorithms.

\paragraph{Evaluation.} Our evaluation methodology aligns with the settings of previous works \cite{zhao2020sess,wang20213dioumatch}. Both ScanNet and SUN RGB-D undergo evaluations using different proportions of randomly sampled labeled data from all the training data. Every evaluation ensures representation of all classes in the labeled data.
In terms of method evaluation, a VoteNet-based 3DIoUMatch is applied on both ScanNet and SUN RGB-D. The metrics reported for these experiments include mean average precision with 3D IoU thresholds of 0.25 (mAP@0.25) and 0.5 (mAP@0.5). Furthermore, performance comparisons with previous state-of-the-art methods, such as SESS and 3DIoUMatch, have proven that innovative approaches can surpass existing methods across all ratios of labeled data and on both datasets.

\subsection{Implementation details}
\textbf{Training scheme.} All experiments could be conducted on a single Nvidia A100 40GB. During the training phase, we utilize an Adam optimizer with an initial learning rate of 0.004 for 1,000 epochs, with the learning rate being decayed at epoch 400, 600, 800, and 900.
The backbone detector used in 3DIoUMatch is an IoU-aware VoteNet, which is a VoteNet model equipped with a 3D IoU estimation module. This backbone detector is also utilized in the experiments. The pretrained model is used not only for initializing the training stage but also for generating predicted proposals for unlabeled data for collision detection. At the training stage, a batch size of {4 + 8} is used, where 4 represents the number of labeled frames, and 8 represents the number of unlabeled frames. During the inference stage, the student IoU-aware VoteNet is used to process input data, demonstrating the effectiveness and versatility of the models and methods employed across both ScanNet and SUN RGB-D datasets.

\begin{table*}[t]
\centering
    \resizebox{\textwidth}{!}{
        \begin{tabular}{c|cc|cc||cc|cc|cc||cc|cc|cc}
\hline\hline
\multirow{2}{*}{ID} & \multicolumn{2}{c|}{Class-Prob.} & \multicolumn{2}{c||}{Geometry.} & \multicolumn{2}{c|}{ScanNet 5\%} & \multicolumn{2}{c|}{ScanNet 10\%} & \multicolumn{2}{c||}{ScanNet 20\%} & \multicolumn{2}{c|}{SUN RGB-D 5\%} & \multicolumn{2}{c|}{SUN RGB-D 10\%} & \multicolumn{2}{c}{SUN RGB-D 20\%} \\ \cline{2-17} 
                    & uni.            & class.         & feat.          & geo.          & mAP@0.25        & mAP@0.5        & mAP@0.25        & mAP@0.5         & mAP@0.25        & mAP@0.5         & mAP@0.25         & mAP@0.5         & mAP@0.25         & mAP@0.5          & mAP@0.25         & mAP@0.5         \\ \hline\hline
(a)                 &                 &                &                &               & 39.44           & 22.19          & 47.53           & 28.63           & 52.70           & 35.33           & 38.72            & 21.31           & 46.02            & 28.88            & 50.39            & 30.71           \\ \hline
(b)                 & \checkmark      &                &                &               & 41.94           & 24.86          & 48.32           & 30.15           & 55.39           & 37.61           & 39.79            & 22.6            & 48.57            & 31.06            & 51.43            & 33.21           \\ \hline
(c)                 & \checkmark      & \checkmark     &                &               & 42.98           & 25.64          & 50.63           & 31.72           & 56.01           & 39.71           & 40.35            & 23.75           & 49.39            & 31.42            & 52.11            & 33.43           \\ \hline
(d)                 & \checkmark      &                & \checkmark     & \checkmark    & 43.40           & 26.88          & 50.82           & 32.60           & 56.87           & 40.57           & 40.89            & 23.9            & 49.56            & 31.94            & 52.95            & 34.05           \\ \hline
(e)                 & \checkmark      & \checkmark     & \checkmark     & \checkmark    & \textbf{44.01}  & \textbf{27.04} & \textbf{51.88}  & \textbf{34.06}  & \textbf{57.01}  & \textbf{40.74}  & \textbf{41.51}   & \textbf{24.98}  & \textbf{49.93}   & \textbf{32.48}   & \textbf{53.26}   & \textbf{35.01}  \\ \hline\hline
(f)                 &                 &                &                &               & 39.44           & 22.19          & 47.53           & 28.63           & 52.70           & 35.33           & 38.72            & 21.31           & 46.02            & 28.88            & 50.39            & 30.71           \\ \hline
(g)                 &                 &                & \checkmark     &               & 39.96           & 22.32          & 48.59           & 29.34           & 53.62           & 35.86           & 39.19            & 21.86           & 46.89            & 29.38            & 51.16            & 31.28           \\ \hline
(h)                 &                 &                & \checkmark     & \checkmark    & 41.43           & 24.11          & 49.68           & 31.16           & 54.41           & 37.38           & 39.44            & 22.29           & 47.49            & 30.76            & 52.45            & 33.29           \\ \hline
(i)                 & \checkmark      & \checkmark     & \checkmark     &               & 43.12           & 25.90          & 50.89           & 32.37           & 56.50           & 40.36           & 40.79            & 24.27           & 49.41            & 31.99            & 52.78            & 34.32           \\ \hline
(j)                 & \checkmark      & \checkmark     & \checkmark     & \checkmark    & \textbf{44.01}  & \textbf{27.04} & \textbf{51.88}  & \textbf{34.06}  & \textbf{57.01}  & \textbf{40.74}  & \textbf{41.51}   & \textbf{24.98}  & \textbf{49.93}   & \textbf{32.48}   & \textbf{53.26}   & \textbf{35.01}  \\ \hline\hline
\end{tabular}
    }
    \vspace{-0.1in}
    \captionsetup{type=table}
\caption{\small{\textbf{Additional ablation studies on ScanNet and SUN-RGBD val sets with various label ratios for training.} These comprehensive ablation experiments conducted on the proposed DPKE can validate the independent effectiveness of each individual module.}}
    \label{tab:more_ablation}
\vspace{0.2cm}
\end{table*}
\begin{table}[]
    \centering
    \scriptsize
    \begin{tabular}{l|c|c} \toprule
        Method & Memory (GB)& Time (hour) \\ \midrule
        3DIoUMatch & 10.2 & 5.75 \\
        Ours with non-optimized $\mathcal{W}_{\text{geometry}}$ & $\geq$40 & - \\
        Ours with efficient $\mathcal{W}_{\text{geometry}}$ ($M_0$=1000) & 13.5 & 8.38 \\
        Ours with efficient $\mathcal{W}_{\text{geometry}}$ ($M_0$=500)  & 11.4 & 6.5 \\ \bottomrule
    \end{tabular}
    \vspace{-0.1in}
    \caption{\small{Cost comparison between 3DIoUMatch, our method with \textit{non-optimized} geometry-aware weight $\mathcal{W}_{\text{geometry}}$ computation, and our method with efficient $\mathcal{W}_{\text{geometry}}$ computation. The GPU memory cost and the training time are obtained when training ScanNet.}}
    \label{tab:efficient}
        \vspace{-0.1in}
\end{table}

\paragraph{The sub sampling strategy.}
However, directly calculating $\mathcal{W}_{\text{chamfer}}^{s, t}$ may consume too many computational resources because we have no limitations on the number of points in $P^t$ and $P^s$. To solve this problem, we will fix the sampling number to be $M_0=500$. If $M_t$ or $M_s$ is less than $M_0$, we randomly re-sample points from existing points and fill them into the point clouds until the total number reaches $M_0$. If $M_t$ or $M_s$ is larger than $M_0$, Farthest Point Sampling (FPS) is adopted to sample $M_0$ points from existing point clouds to ensure that the overall shape of the original point cloud is maintained. If there is no point in either the $P^t$ or $P^s$ of a proposal pair, we will set the corresponding weight $W_i$ to be the smallest value $W_{\text{threshold}}$.

\paragraph{Hyper-parameters.} We follow all the hyper-parameters in 3DIoUMatch. Here we give the extra hyper-parameters in our methods. The $N$ mentioned for the epochs when sampling from proposal bank is set as 600. The $\delta$ in the feature matching loss is set as 1. The $o^t$ utilized in our geometry-aware feature matching loss is set as 0.6 for both ScanNet and SUN RGB-D.

\section{Additional experiments}\label{sec:additional}

\textbf{More ablation studies with different ratios.} Thank you for your valuable suggestion. We have included additional ablation results in Tab.~\ref{tab:more_ablation}, considering additional ratios of labeled data (5\%, 10\%, and 20\%). These supplementary ablation studies underscore the effectiveness of each of our contributions across a range of label ratios.

\textbf{The adaptability towards different threshold.} We change the thresholds of filters in 3DIoUMatch and our proposed method. The results on ScanNet and SUN RGB-D are shown in Figure~\ref{fig:threshold}. On different datasets with different ratios, the mAP curves show similar tendancy. When the threshold becomes smaller, the performances of 3DIoUMatch dramatically becomes smaller while the performances of our proposed geometry-aware feature matching loss does not change much. The performances of our proposed geometry-aware feature matching loss and 3DIoUMatch become closer because the samples that are supervised by the geometry-aware feature matching loss become less. It is obvious that our method is more robust towards the change of the threshold.

\paragraph{Comparison between 3DIoUMatch and the method simply combining SESS and 3DIoUMatch.} Similar with SESS's consistency loss, our geometry-aware feature matching loss is a kind of soft supervision compared with 3DIoUMatch's loss. However, the differences is that our proposed geometry-aware feature matching loss will not conflict with 3DIoUMatch's loss. We do experiments on both ScanNet and SUN RGB-D to verify it. The results are shown in Table~\ref{tab:comparison}. Adding the consistency loss from SESS to 3DIoUMatch will even harm the performances on both datasets, while our DPKE can consistently improve the performances based on 3DIoUMatch.

\textbf{Analyses about the computational resources introduced by the geometric-aware weight.} 
We have added the analysis of the memory and time cost in Tab.~\ref{tab:efficient}. When the efficient computation for $\mathcal{W}_{\text{geometry}}$ is disabled (2nd row), certain instances would arise in which the number of points within object proposals becomes excessively large, exceeding the GPU memory capacity (40GB). This leads to program crashes, making it challenging to estimate the time cost. When employing our efficient method to compute $\mathcal{W}_{\text{geometry}}$, we introduce a hyperparameter $M_0$ to govern the number of points sampled using FPS. Tab.~\ref{tab:efficient} shows that by selecting a relatively small value for $M_0$, the memory cost and training time of our method can be significantly reduced, resulting in costs similar to those of 3DIoUMatch.



\end{appendix}

\end{document}